\documentclass[11pt,a4paper]{article}

\usepackage{ona}
\usepackage{times}
\usepackage{latexsym}
\usepackage{url}

\usepackage{color,graphicx} % for dvips(k)

\title{An End-to-End Khmer Optical Character Recognition using Sequence-to-Sequence with Attention}
\author{Rina Buoy\affilA   \and Sokchea Kor \affilB \and Nguonly Taing\affilA \\
	\affilA Techo Startup Center (TSC) \\
	\affilB Royal University of Phnom Penh (RUPP) \\
	{\tt \{rina.buoy,nguonly.taing\}@techostartup.center} \\ {\tt kor.sokchea@rupp.edu.kh}}

\summary{
This paper presents an end-to-end deep convolutional recurrent neural network solution for Khmer optical character recognition (OCR) task. The proposed solution uses a sequence-to-sequence (Seq2Seq) architecture with attention mechanism. The encoder extracts visual features from an input text-line image via layers of residual convolutional blocks and a layer of gated recurrent units (GRU). The features are encoded in a single context vector and a sequence of hidden states which are fed to the decoder for decoding one character at a time until a special end-of-sentence (EOS) token is reached. The attention mechanism allows the decoder network to adaptively select parts of the input image while predicting a target character. The Seq2Seq Khmer OCR network was trained on a large collection of computer-generated text-line images for seven common Khmer fonts. The proposed model's performance outperformed the state-of-art Tesseract OCR engine for Khmer language on the 3000-images test set by achieving a character error rate (CER) of 1\% vs 3\%. 
}

\begin{document}
\maketitle


\section{Introduction}
\subsection{Optical Character Regognition (OCR)}

One of the artificial intelligence (AI) paradigms is to develop a machine that can mimic the ability of human recognition. In term of visual perception and understanding text, the computer is still at the infancy level, compared with human \cite{bayram}. 

To make a text machine-readable, it can be either converted manually or digitally extracted by mean of Optical Character Recognition (OCR), from the digital image of the document \cite{9151144}.

OCR is the science of extracting analyzable and editable data from the scanned documents or images. The  OCR technology has been evolving over the last 8 decades. The early phase of OCR development was mainly contributed by large tech players. The recent advancement of artificial intelligence, particularly deep learning has allowed researchers from various spectrum to devise OCR algorithms that can achieve higher accuracy levels \cite{9151144}.

Although OCR technology for English and other high-resource languages has been developed over the last 8 decades  \cite{9151144}, the earlier OCR work on the Khmer language was around the year 2005. 

\subsection{Khmer Writing System}
Khmer (KHM) is the official language of the kingdom of Cambodia. The Khmer script is used in the writing system of Khmer and other minority languages such Kuay, Tampuan, Jarai Krung, Brao and Kravet. Khmer language and writing system were hugely influenced by Pali and Sanskrit in early history \cite{buoy2021joint,MSok}. 

Unlike Latin-based languages, the Khmer language has a complex writing system. One or two consonants can be stacked below an initial consonant using the alternate form (aka Coeng - foot in English) to form a consonant cluster \cite{buoy2021joint}. Khmer writing also uses diacritical signs which are placed above a consonant. Dependent vowels cannot stay alone by themselves and have to be attached to an initial consonant \cite{buoy2020}. Orthographically, a dependent vowel can be placed to the left, right, above, below, or around a base consonant \cite{ding_nova_2018}. 

\begin{figure}[ht]
	\begin{center}
		\includegraphics[width=7cm]{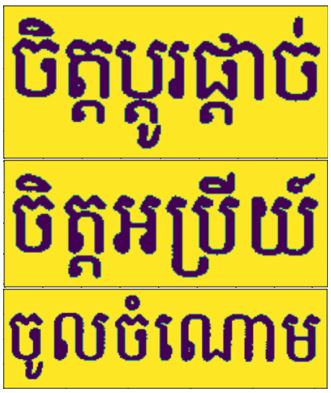}
		\caption{A few examples of Khmer words illustrating complex glyph layout}
		\label{examplewords}
	\end{center}
\end{figure}

Therefore, Khmer scripts require complex rendering layout which is not the case of  Latin-based writing systems. A few examples of Khmer words are selected for illustration purpose as given in \figref{examplewords}.

A complete Khmer OCR system needs to recognize all characters, given the complexity of Khmer writing. 

\section{Related Work}

\subsection{OCR for Khmer Language}

One of the early work in Khmer OCR was done by \cite{chey_et_al}. The proposed method was a variant of instance-based classifiers. The authors used wavelet descriptors to extract features (coefficients) from images in the training set and built a template for each character. For a given new input image, a set of wavelet coefficients was extracted and  matched against all the training templates. The input image was then assigned to the class with the smallest Euclidean distance.

\cite{pan_cambodia}, from PAN Localization Cambodia team, proposed a complete workflow for recognizing Khmer text.  The four-steps workflow included pre-processing, segmentation, recognition and mapping. The pre-processing step included line separation and character block segmentation. Blocks of characters were then segmented into atomic shapes   namely: Main Body, SuperScript, SubScript, CCDown, and CC (Complex Character). Discrete cosine transform was used to extract features from the atomic shapes for classification task. The recognized shapes were finally mapped to produce valid Khmer text. The average recognition rate was reported to be 96.34%. 

\cite{hann} applied artificial neural network in recognizing Khmer characters. The approach applied to individual character recognition. The authors proposed two-steps recognition pipeline. An input image (20 by 20 pixels) was first passed into a self-organizing network which grouped the input image into one of the 9 classes. Each class had one multilayer neural network which classified the input image to one of 82 Khmer characters including consonant, vowels and numbers.

\cite{dona_2017} proposed a convolution neural network (CNN) classifier in recognizing ancient Khmer characters from palm leaf manuscript. The proposed CNN architecture was applied to individual character recognition and is composed of 3 convolutional blocks and a linear classifier. The classifier output a vector of 106 elements representing character classes. The accuracy on the test set was reported around 95.96\%. 

\cite{pong} applied support vector machine SVM algorithm to recognize Khmer characters. The proposed pipeline was composed of four steps - character segmentation, feature extract, classification and character re-assembling. The reported character classification accuracy was about 98\% for various font sizes.

\cite{bayram} experimented both multilayer  and convolutional neural networks to recognize Khmer consonants. A CNN-based model was compared against artificial neural network (ANN)-based classifier with full feature set and ANN-based classifier with reduced feature set. The CNN model achieved up to 94.85\% average accuracy. 

\cite{kim} fined tune a pre-trained Tesseract OCR engine for khmer Unicode and legacy Lemon fonts.  Tesseract is an end-to-end multilingual OCR engine. Tesseract uses deep convolutional recurrent neural network architecture with connectionist temporal classification (CTC) loss. Tesseract learns feature representation automatically via one convolutional layer followed by multiple stacked LSTMs\cite{liebl2020accuracy}. Tesseract can recognize a text-line image. \cite{kim} reported the accuracy of 90\% on the trained fonts.

\subsection{Encoder-Decoder Networks}

Although deep neural networks (DNNs) are very good models in computer vision or natural language processing, DNNs are not able to handle inputs and targets of variable length. This limitation prevents DNNs from being applied to certain tasks such as speech recognition and machine translation, in which sequence lengths are not fixed \cite{sutskever2014sequence}. 

Recurrent neural networks (RNNs), on the other hand, can encode an input sequence of unknown length to produce a fixed-dimension representation of the input sequence. Thus, \cite{sutskever2014sequence} proposed an RNN known as called encoder to encode the input sequence of variable length to a fixed-dimension vector representation, commonly known as a context vector. The context vector was passed to another RNN known as decoder to extract the information and generate the output sequence any length. 

\cite{sutskever2014sequence} applied the encoder-decoder network also known as Seq2Seq to the English-French machine translation task. \cite{sutskever2014sequence} achieved a BLEU score of 34.81 which was, at that time, the best result by neural machine translation. \cite{sutskever2014sequence} made the observation that the the Long Short-Term Memory (LSTM) architecture performed robustly even on long sequences although previous study suggested otherwise. 

\cite{bahdanau2014neural} suggested that using a single fix-dimension context is a bottleneck for the encoder-decoder network. To remove this bottleneck, \cite{bahdanau2014neural} proposed the attention mechanism to allow the decoder to selectively use only parts of the input sequence that are useful for predicting a target word. 

The key distinction between  \cite{sutskever2014sequence} and \cite{bahdanau2014neural} is that \cite{sutskever2014sequence} proposed  a single fixed-length vector for the whole input sentence, while \cite{bahdanau2014neural} proposed to encode the input sequence as a sequence of vectors and the decoder chooses a subset of these vectors adaptively via attention mechanism. 
while decoding the translation.

The experimental results from \cite{bahdanau2014neural} suggested that the proposed approach outperformed the conventional encoder–decoder significantly on the English-to-French translation task.

OCR task can be viewed as another machine translation task. The input to the OCR systems is an input image of a text-line with unknown length. The output of the OCR systems is a character sequence of unknown length. Therefore, the solutions to machine translation task can also be applied to OCR task. 

\cite{sahu2015sequence} proposed an end-to-end encoder-decoder network for recognizing printed text. \cite{sahu2015sequence} used long short-term memory (LSTM) as both word image reader (encoder) and word image predictor (decoder). \cite{sahu2015sequence} reported the label rate of 0.84 \% on the annotated English word images. 

\cite{safir2021endtoend} proposed an end-to-end convolutional recurrent neural network with connectionist temporal classification (CTC) loss for Bengali handwritten words. \cite{safir2021endtoend} experimented with various CNN feature extractors such as DenseNet, Xception, NASNet, and MobileNet and different recurrent neural networks such as LSTM and GRU. \cite{safir2021endtoend} reported 0.091 character error rate (CER) and 0.273 word error rate (WER) performed using DenseNet121 model with GRU recurrent layer.

In this paper, we attempt to approach the Khmer OCR task by using an encoder-decoder network. We propose an end-to-end, attention-based encoder-decoder network (Seq2Seq). The encoder is a convolutional recurrent neural network with skip (residual) connections while the decoder is a recurrent neural network and a linear classifier. During decoding, the attentive decoder uses an attention mechanism to search for the relevant encoder outputs. Gated Recurrent Unit (GRU) is used in both the decoder and encoder. The entire OCR network can read a text-line image of arbitrary length and produces a character sequence of unknown length. Being an end-to-end solution, the proposed solution does not need any pre-processing, feature extraction and post-processing steps.

The overall contributions of this work to Khmer OCR are as follows:
\begin{itemize}
	\item	We introduce one of the first end-to-end solutions to Khmer OCR. 
	\item	The proposed solution outperforms the current state-of-art Tesseract engine for Khmer language. 
	
\end{itemize}

\section{End-to-End Khmer OCR Model}

We present the end-to-end Khmer OCR model using a Seq2Seq network with an attention mechanism. The end-to-end OCR model does not require pre-processing steps such as character separation and post-processing steps such as character mapping or assembling. 	

The proposed Seq2Seq network takes a text-line input image via the encoder network. The encoder network passes the encoded information via a context vector to the decoder network to decode one character at a time until end-of-sentence token is reached. There is no limit on the number of characters in the input text-line image that the encoder can read and the decoder can produce. The attention mechanism allows the network performs robustly even with an input image with very long text-line. 

\begin{figure*}
	\begin{center}
		\centering
		\includegraphics[width=\textwidth]{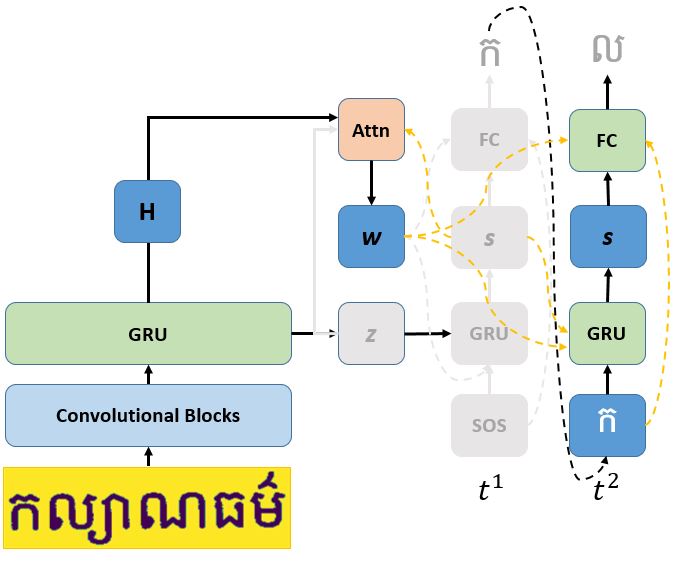}
		\caption{The end-to-end, attention-based Seq2Seq network for Khmer OCR } 
		\label{encoderdecoder}
	\end{center}
\end{figure*}

The high-level architecture of the end-to-end, attention-based Seq2Seq network for Khmer OCR is given in \figref{encoderdecoder}. \figref{encoderdecoder} shows the decoding process of the second timestep while the decoding process of the timestep is being grayed out.  Detailed description of the encoder, decoder and attention is provided in the subsequent sections. 

\subsection{Encoder Network}
\begin{figure*}
	\begin{center}
		\centering
		\includegraphics[width=\textwidth]{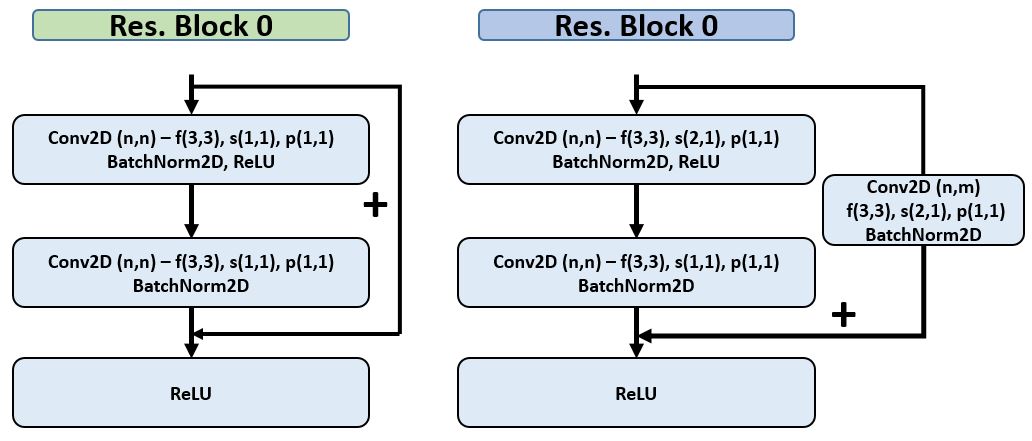}
		\caption{The residual blocks - with and without downsampling} 
		\label{residualblock}
	\end{center}
\end{figure*}

The encoder network is composed of a series of convolutional blocks with skip connections and a layer of GRU units.  The encoder network is composed of the followings:

\begin{enumerate}
	\item A convolutional layer with 3x3 filter, 1x1 padding and 1x1 stride. This convolutional layer has input channel of 1 and output channel of 16. Batch normalization and ReLU are applied to the resulting feature maps. 
	\begin{equation}
		O_{conv1} = Conv2D(img)
	\end{equation}
where:
	\begin{itemize}
		\item	$O_{conv1}$: are the feature maps.
		\item	$img$: is the text-line input image.
	\end{itemize}
	\item The first convolution layer is followed by 3 layers of the residual blocks. Each layer has 2 residual blocks. There are two types of the residual blocks:
	\begin{itemize}
		\item The residual block without downsampling is shown in \figref{residualblock}. There is no change to the dimension of the feature maps within this residual block. 
		\item The residual block with downsampling is shown in \figref{residualblock}. In the first convolution of the residual block, 2x1 stride is applied and thus, the height of the resulting feature maps is reduced by half. Therefore, the input feature maps to the residual block must be downsampled before being adding to the output feature maps from the last convolution in the residual block.

	\end{itemize}

		\begin{equation}
	O_{resnet} = ResLayers(O_{conv1})
\end{equation}
\begin{itemize}
	\item	$O_{resnet} $: are the feature maps.
	\item	$ResLayers$: are the 3 layers of residual blocks.
	
\end{itemize}

	The input feature maps to the first layer of residual blocks have 16 channels and the number of channels is maintained across the first layer. The second and third layer of the residual blocks increase the depth of the feature maps twice in each layer. That means after the third layer of the residual blocks, the feature maps have 64 channels.
	
	\item Average pooling with 8x8 filter and 8x6 stride is applied to the feature maps from the last layer of residual blocks, followed by a 2D dropout layer. The resulting feature maps are reshaped into a width-by-channel format by flattening the height into the channel dimension. 
	
	\begin{equation}
		O_{avgpooling} = AvgPooling2D(O_{resnet} )
	\end{equation}
	\begin{equation}
		O_{dropout2D} = Dropout2D(O_{avgpooling})
	\end{equation}

	\item Finally, the width-by-channel input is fed to a layer of bidirectional GRU units. The GRU recurrent network produces encoder hidden states  - $H$ = \{$h_1$,$h_2$,...,$h_T$\}. An encoder hidden state, $h_t$ ($[h_t^\rightarrow; h_{T-t-1}^\leftarrow]$) is a concatenated forward and backward hidden state, at a timestep, $t$ in the input sequence (i.e. width of the final feature maps). The last hidden state of the forward GRU layer, $h_T^\rightarrow$ is fed to a dropout layer and then, a linear layer followed by a $tanh$ activation function to produce a hidden state vector $z$ as its initial hidden state, $s_0$ for the decoder GRU network. 
	
	\begin{equation}
		H,h_T^\rightarrow = EncoderGRU(O_{dropout2D})
	\end{equation}

	\begin{equation}
	z = Tanh(LinearLayer(dropout(h_T^\rightarrow)))
\end{equation}

\end{enumerate}

In a nutshell, the encoder network take an input image with a text-line and returns  a sequence of hidden states, $H$ and the decoder's initial hidden state,  $s_0$ which is $z$.

\begin{figure}[ht]
	\begin{center}
		\includegraphics[width=7cm]{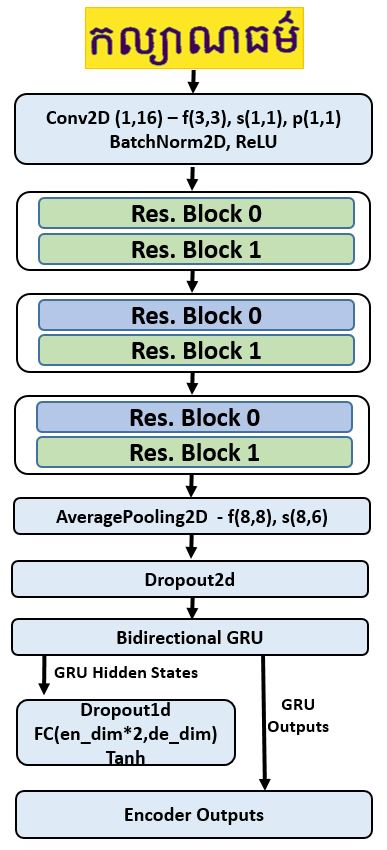}
		\caption{The architecture of the encoder network}
		\label{encoder}
	\end{center}
\end{figure}

\subsection{Attention}

At each decoding step $i$, the decoder needs to search the relevant information from the encoder's hidden states adaptively. This can be done by calculating a weighted average context vector, $c_i$,  over the encoder's hidden states \cite{bahdanau2014neural}. 
\begin{equation}
	c_i = \sum_{j=1}^{T_x} \alpha_{ij}*h_j 
\end{equation}
where:
\begin{itemize}
	\item $T_x$ is the length of the input sequence. 
	\item $c_i$ is the weighted average context vector at decoding step $i$.
	\item $\alpha_{ij}$ is the attention weight for the encoder's hidden state at time  $j$ and the decoder's previous hidden state at time $i$.
	\item $h_j$ is the encoder's hidden state at time  $j$.
\end{itemize}

$\alpha_{ij}$ can be computed using the below formula \cite{bahdanau2014neural}.

\begin{equation}
	\alpha_{ij} = \frac{e^{e_{ij}}}{ \sum_{k=1}^{T_x} e^{e_{ik}}}
\end{equation}

\begin{equation}
	e_{ij} = v^{T}_a tanh(W_as_{i-1} + U_ah_j)
\end{equation}

where:
\begin{itemize}
	\item $v_a$, $W_a$, and $U_a$ are weight matrices.
	\item $s_{i-1}$ is the decoder's previous hidden state.
\end{itemize}

\subsection{Decoder Network}

The decoder network is a single layer GRU network followed by a linear classifier. The inputs to the GRU network are:
\begin{itemize}
 \item $d(y_t)$, the previous decoded target that is one-hot encoded. For $t=1$, $y_0$  is a special start-of-sentence (SOS) token.
 \item $s_{t-1}$, the decoder's previous hidden state.
 \item $c_{t}$, the weighted average context vector at decoding step $t$.
\end{itemize}

$d(y_t)$ and  $c_{t}$ are concatenated as a single input vector to the GRU layer. Together with  the previous hidden state, $s_{t-1}$, the GRU layer produces a new hidden state, $s_{t}$. 

\begin{equation}
	s_{t} = DecoderGRU([d(y_t),c_{t}], s_{t-1})
\end{equation}

Next,$s_{t}$, $d(y_t)$ and  $c_{t}$ are concatenated as a single input vector to the linear classifier to predict the target, $y_{t+1}$. 

\begin{equation}
	y_{t+1} = LinearClassifer([s_{t},c_{t},d(y_t)])
\end{equation}

The decoding process is repeated until $y_{t+1}$ is a special end-of-sentence (EOS) token

\section{Experiments}
The proposed solution was applied to Khmer OCR task. Each input image contains a text-line of Khmer words, phrases or sentences from Khmer Asian Language Treebank (ALT) \footnote{https://www2.nict.go.jp/astrec-att/member/mutiyama/ALT/} . The dataset was generated by using text2image \footnote{developed by Tesseract team} tool for a number of common Khmer fonts. The rendered images were augmented with speckle noise, dilation/erosion and rotation. The proposed solution was benchmarked against the state-of-art OCR tool for Khmer language which is Tesseract. 
\subsection{Dataset Details}
\begin{figure*}
	\begin{center}
		\centering
		\includegraphics[width=\textwidth]{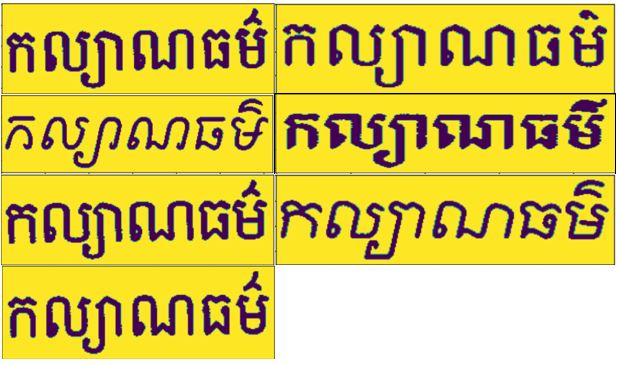}
		\caption{Different input images of a Khmer word for the different fonts } 
		\label{imagefonts}
	\end{center}
\end{figure*}

The train dataset consists of 92,213 words, phrases or sentences. The text2image was used to generate the corresponding images for a number of Khmer fonts including  OS, OS Siemreap, OS Battambang, OS Bokor, OS Freehand, and OS Fasthand. The images for different fonts for the same word is given in \figref{imagefonts}. The test dataset consists of 3000 images. 

The text2image generates images with variable width and height, depending on the word or phrase or sentence length and the presence of subscript consonants, vowels and  diacritics. Therefore, the input images are scaled to a common height of 64 pixels that givens the final height of 2 after the convolutional layers in the encoder network. According to \cite{liebl2020accuracy}, Tesseract uses a common height of 48 pixels. 

\subsection{Training Details}

The following hyper-parameters were used to train the proposed Seq2Se2 model. 

\begin{itemize}
	\item $outdim_{decoder}$ is the output dimension of the linear classifier in the decoder network. $outdim_{decoder}$ is equal to the number of Khmer characters including vowels (dependent + independent), consonants, diacritics, and numbers.
	\item $hiddendim_{encoder}$  is the hidden dimension of the GRU layer in the encoder. 300 was used.
	\item $hiddendim_{decoder}$  is the hidden dimension of the GRU layer in the decoder. 300 was used.
	\item $Dropout_{encoder}$  is the dropout probability in the encoder network. 0.2 was used.
	\item $Dropout_{decoder}$  is the dropout probability in the decoder network. 0.5 was used.
	\item $Optimizer$  is Adam optimizer with initial learning rate (LR) of 0.001.
	\item $LR\ schedule$: the learning rate was decreased by half every 10 epochs.
	\item $Batch\ size$ is the number of input images loaded per optimization iteration. 32 was used. 
	\item $Teacher\ Forcing\ Ratio$ is the probability of using the actual target for $y_t$ instead of the decoder's prediction at time $t$. The value of 0.5 was used.
	\item $Cross\ Entropy$ loss function was used.
\end{itemize}

The proposed Seq2Seq model was implemented in PyTorch \footnote{adapted from https://github.com/bentrevett/pytorch-seq2seq} and the model was trained on Google Colab virtual machine. 
\section{Results and Discussions}
\subsection{Evaluation Metrics - CER and WER}
The trained model was evaluated on the test set of 3000 images. The model achieved the perplexity of 1.037 and average cross-entropy loss of 0.037. Character and word error rate were also computed. The character error rate (CER) is the ratio of unrecognized characters over the total number of characters in the target words. The word error rate(WER) is defined similarly. CER was computed using Levenshtein distance. 

The CER is defined as below \cite{safir2021endtoend}:

\begin{equation}
	CER = \frac{S+I+D}{N}
\end{equation}
where:
\begin{itemize}
	\item $S$ is number of substitutions.
	\item $I$ is number of inserts.
	\item $D$ is number of deletes.
	\item $N$ is number of ground-truth characters .
\end{itemize}

The WER is defined as below:

\begin{equation}
	WER = \frac{ count\ of\ incorrect\ samples}{count\ of\ samples}
\end{equation}

In this study, WER is a conservative error metric since each sample image can at minimum a word and at maximum a sentence.

The model achieved a CER of 1\%  and 9\% WER on the test set while Tesseract gave 3\% CER and 26\% WER.

\subsection{Error Analysis}

\begin{figure*}
	\begin{center}
		\includegraphics[width=7cm]{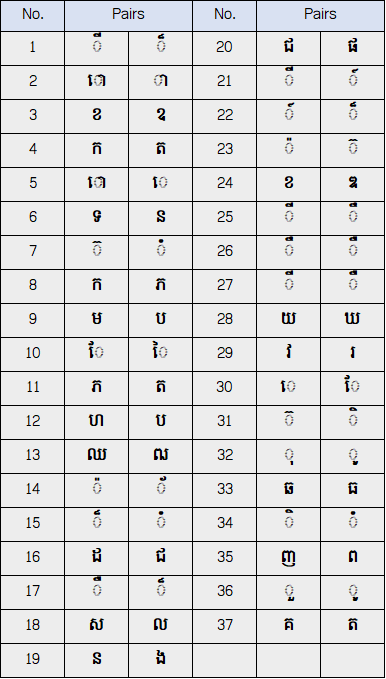}
		\caption{Pairs of Khmer characters with similar glyphs}
		\label{similarpairs}
	\end{center}
\end{figure*}

We investigated the errors produced by Tesseract and the proposed model. The following observations were made:

\begin{itemize}
	\item Some of the erroneous cases by Tesseract were due to that Tesseract was likely to produce symbols such as @, \#, \textbackslash \  that do not present in the ground truth. This is not the case of the proposed model since the proposed model outputs only Khmer characters and symbols only.
	\item Some of the erroneous cases by the proposed model were due to that the model was likely to produce repeated characters before EOS token was reached. This is not the case of Tesseract since the Tesseract does not have separate encoder and decoder networks.
	\item Most of the erroneous cases by Tesseract and the proposed model were due to that some Khmer characters have similar glyphs. The pairs of Khmer characters with similar glyphs are given in \figref{similarpairs}. These pairs were identified by aligning the erroneous cases and their corresponding ground truths. 

\end{itemize}

\subsection{Visualizing Attention}

\begin{figure*}
	\begin{center}
		\centering
		\includegraphics[width=\textwidth]{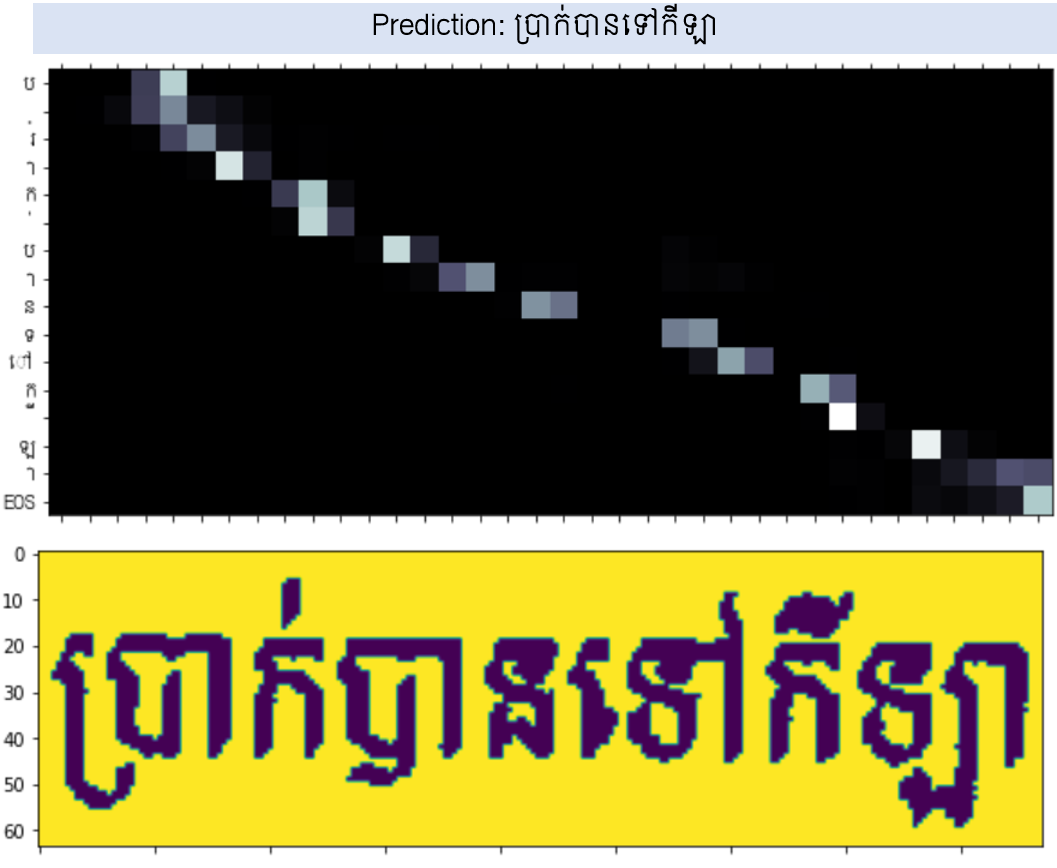}
		\caption{Example 1 - Input image and attention heatmap} 
		\label{example1}
	\end{center}
\end{figure*}

\begin{figure*}
	\begin{center}
		\centering
		\includegraphics[width=\textwidth]{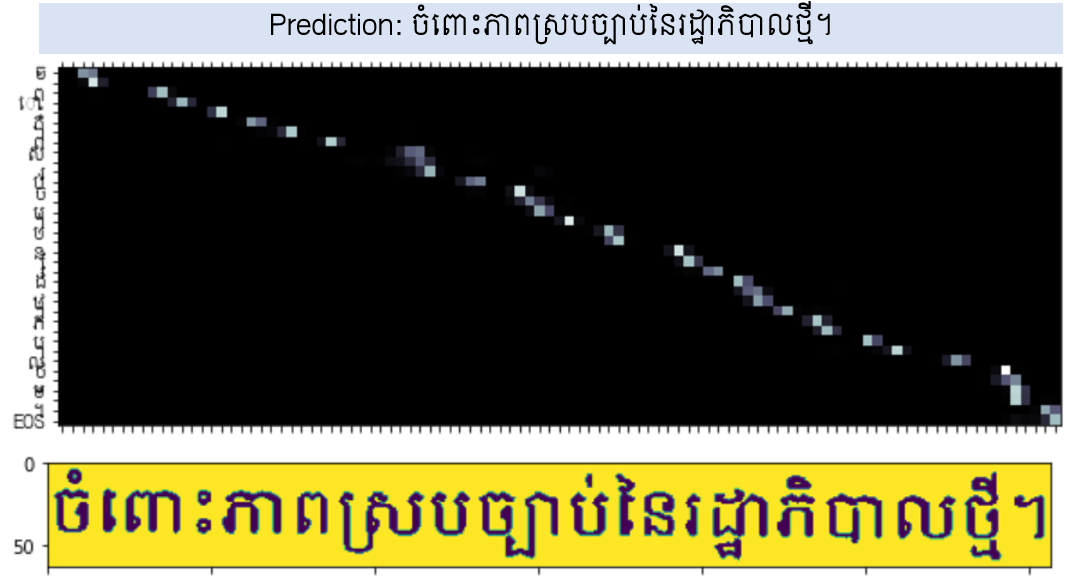}
		\caption{Example 2 - Input image and attention heatmap} 
		\label{example2}
	\end{center}
\end{figure*}

\begin{figure*}
	\begin{center}
		\centering
		\includegraphics[width=\textwidth]{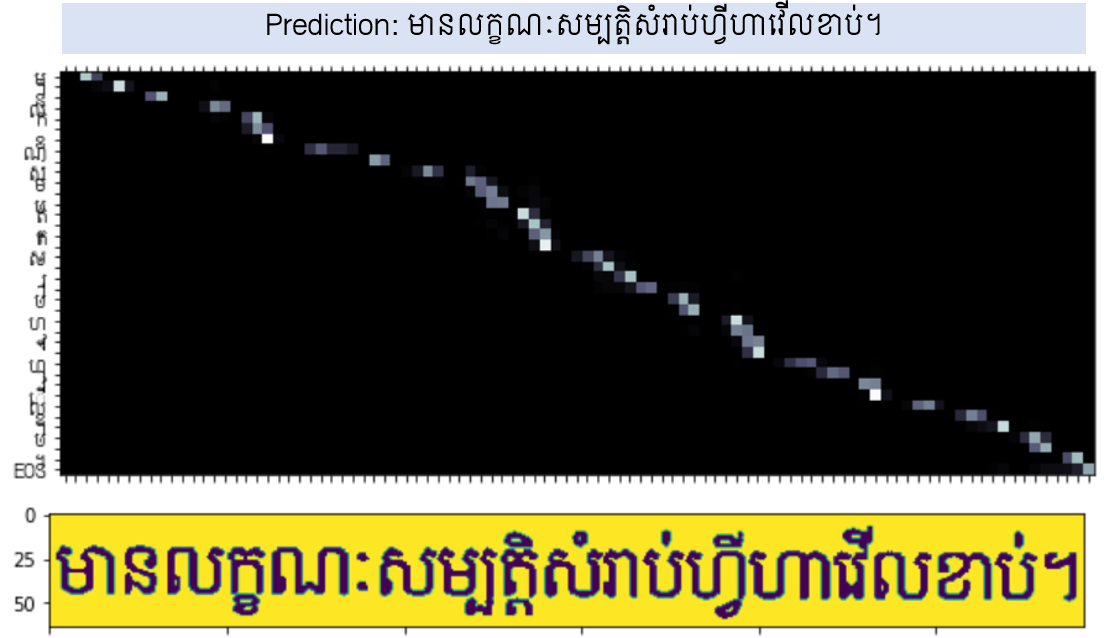}
		\caption{Example 3 - Input image and attention heatmap} 
		\label{example3}
	\end{center}
\end{figure*}

\figref{example1}, \figref{example2} and \figref{example3} are the input images along with the attention heatmaps produced by the model. The attention heatmaps show parts of the input images at which the decoder pays attention at each step. 

\section{Printed Khmer Text Extraction Application}
\subsection{PDF Text Extraction}
\begin{figure*}
	\begin{center}
		\centering
		\includegraphics[width=\textwidth]{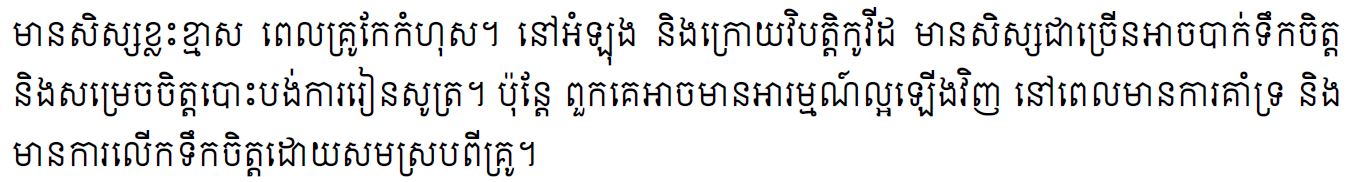}
		\caption{Sample Khmer text in PDF format} 
		\label{khmertextpdf}
	\end{center}
\end{figure*}

The trained model was used to extract text from a PDF document. The sample PDF text is given in \figref{khmertextpdf} \footnote{obtained from http://krou.moeys.gov.kh/}.

Since the trained model accepts input as a text-line image. The following image processing steps were required.

\begin{figure*}
	\begin{center}
		\centering
		\includegraphics[width=\textwidth]{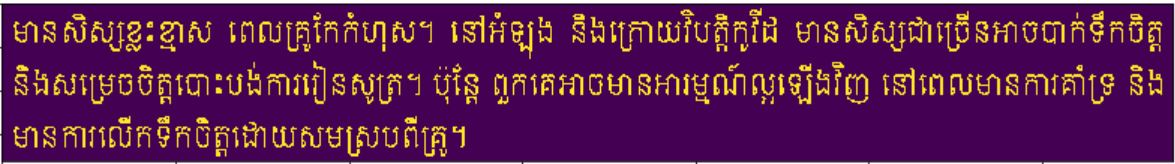}
		\caption{After thresholding} 
		\label{binarizedimg}
	\end{center}
\end{figure*}

\begin{figure*}
	\begin{center}
		\centering
		\includegraphics[width=\textwidth]{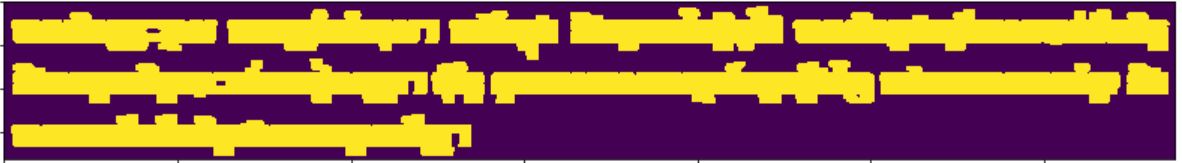}
		\caption{After dilution} 
		\label{dilutedimg}
	\end{center}
\end{figure*}

\begin{figure*}
	\begin{center}
		\centering
		\includegraphics[width=\textwidth]{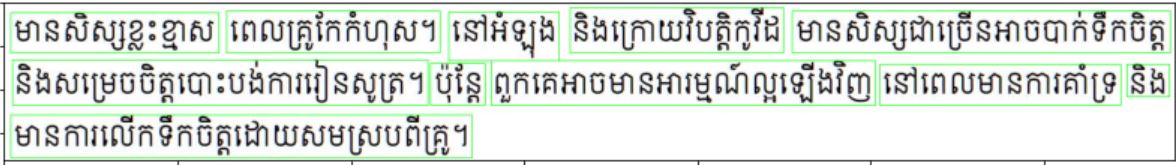}
		\caption{The detected contours} 
		\label{contour}
	\end{center}
\end{figure*}

\begin{itemize}
	\item Thresholding was applied by using Otsu’s thresholding. The resulting image after thresholding is given in \figref{binarizedimg}. 
	\item Dilution was then applied to increase the boundaries of regions of foreground pixels. The resulting image after dilution is given in \figref{dilutedimg}. 
	\item Contour detection was used to detect the borders of the connected characters. The detected contours are shown in \figref{contour}.
\end{itemize}

\begin{figure*}
	\begin{center}
		\centering
		\includegraphics[width=\textwidth]{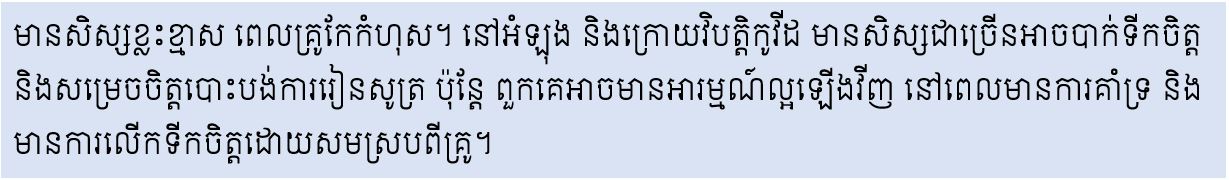}
		\caption{The recognized text, corresponding to \figref{khmertextpdf}} 
		\label{recognizedtext}
	\end{center}
\end{figure*}

Each detected contour was, finally, fed into the trained OCR model to extract the digital text. The recognized text, corresponding to \figref{khmertextpdf}, is given in \figref{recognizedtext}.

\subsection{Camera Text Extraction}
\begin{figure*}
	\begin{center}
		\centering
		\includegraphics[width=\textwidth]{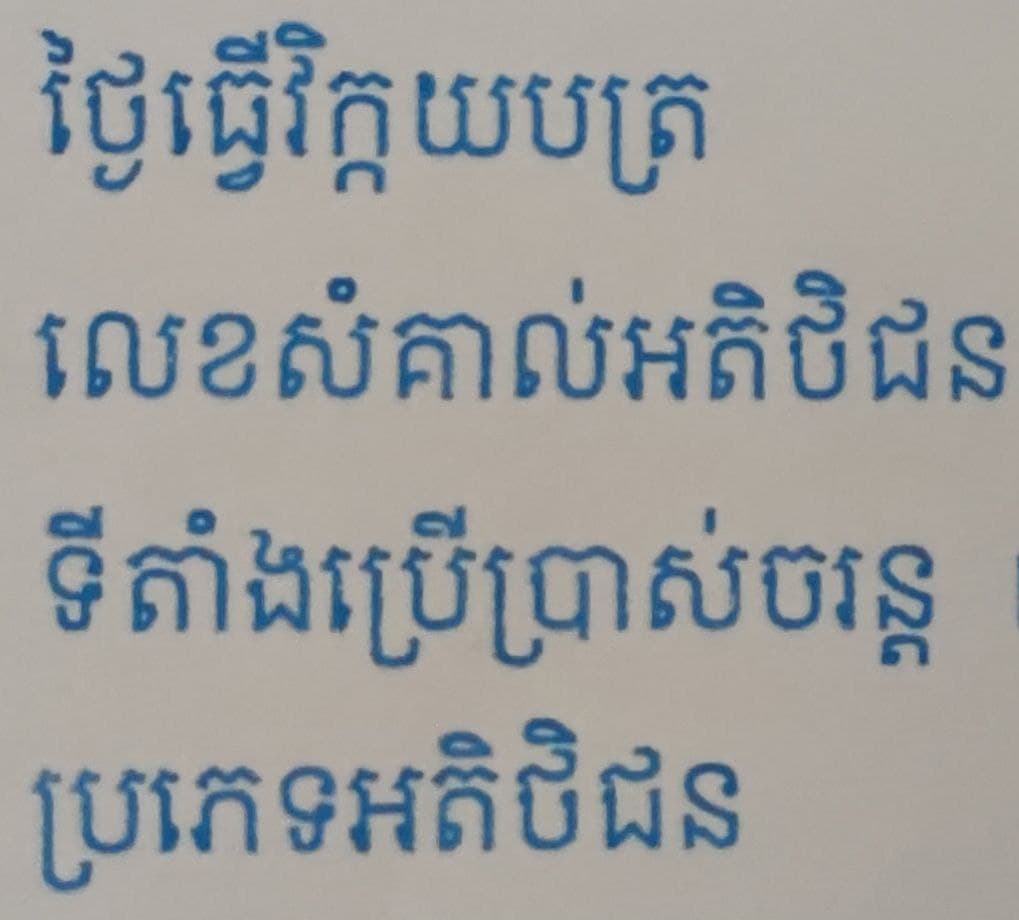}
		\caption{Sample Khmer text from a smart phone's camera} 
		\label{khmertextcamera}
	\end{center}
\end{figure*}

\begin{figure*}
	\begin{center}
		\centering
		\includegraphics[width=\textwidth]{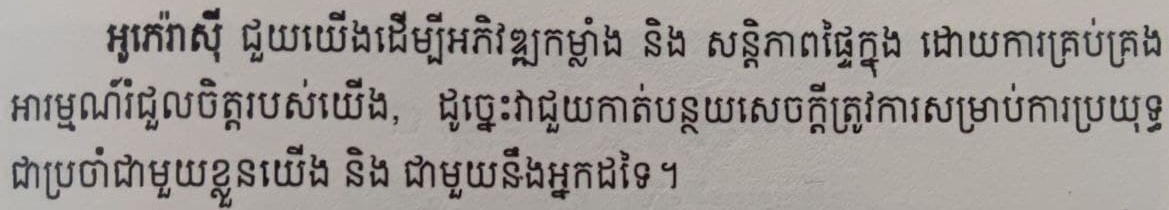}
		\caption{Another ample Khmer text from a smart phone's camera} 
		\label{khmertextcamera2}
	\end{center}
\end{figure*}

We applied the same above extraction processes for text-line images from a smart phone's camera. The camera text-line images are given in \figref{khmertextcamera} and \figref{khmertextcamera2}, illustrating the presence of background and text deformation effects.

\begin{figure*}
	\begin{center}
		\centering
		\includegraphics[width=\textwidth]{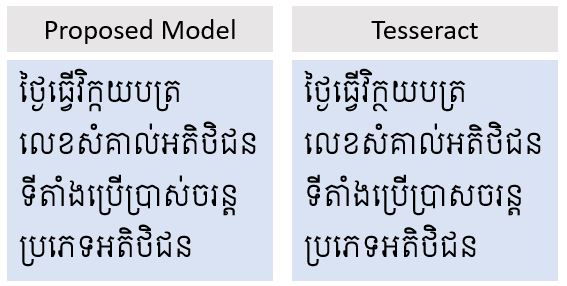}
		\caption{The recognized text by the proposed model and Tesseract} 
		\label{recognizedtextmodel}
	\end{center}
\end{figure*}

\begin{figure*}
	\begin{center}
		\centering
		\includegraphics[width=\textwidth]{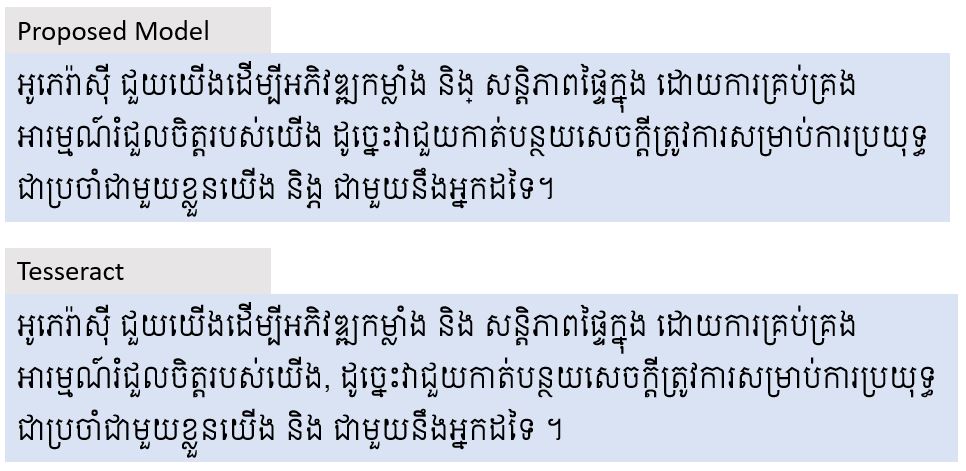}
		\caption{The recognized text by the proposed model and Tesseract} 
		\label{recognizedtextmodel2}
	\end{center}
\end{figure*}

The extracted texts from the proposed model and Tesseract are given in \figref{recognizedtextmodel} and \figref{recognizedtextmodel2}. \figref{recognizedtextmodel} and \figref{recognizedtextmodel2} suggest that both the proposed model and Tesseract are able to recognize text in camera images with acceptable accuracy.

\section {Conclusion and Future Work}

This paper presents an end-to-end Khmer OCR system which utilizes an encoder-decoder (Seq2Seq) network. The encoder is composed of layers of convolutional residual blocks and a layer of GRU units. The decoder consists of a layer of GRU units and a linear classifier. The decoder uses attention mechanism to adaptively select parts of the encoder's outputs that are relevant for predicting the target character.    The proposed model was trained a collection of text-line images generated from ALT dataset by using open-source text2image tool with degrading effects. The experiment results suggested that the proposed solution outperformed the current state-of-art Tesseract engine for Khmer OCR by achieving a CER of 1\% vs 3\% and a WER of 9\% vs 26\% We also present necessary image processing steps to extract text-line images from a PDF document before running the OCR system. 

A robust OCR system must be able to recognize text in presence of text deformations, and noisy background. The system must be also font-invariant. In the future, we would like to train the model on more complex, augmented text-line images and additional Khmer fonts are included.

\bibliographystyle{unsrt}
\bibliography{ona-sample}

\end{document}